\documentclass[final,3p,times,twocolumn]{elsarticle}

\usepackage{amssymb}
\usepackage{multirow}
\usepackage{amsmath}
\usepackage{booktabs}
\usepackage{array}
\usepackage[table]{xcolor}
\usepackage[colorlinks=true,linkcolor=blue,citecolor=blue,urlcolor=blue]{hyperref}

\journal{Nuclear Physics B}

\newcommand{\ie}{\textit{i}.\textit{e}.}
\newcommand{\eg}{\textit{e}.\textit{g}.}

\def\algorithmname{CMTFormer}

\begin{document}

\begin{frontmatter}



\title{CMTFormer: Marrying Transformer with Hierarchical Information Interaction for RGB-Event Object Detection}

\author[label1]{Yu Li}
\ead{liyu23e@nudt.edu.cn}

\author[label2]{Yuenan Hou\corref{cor1}}
\ead{houyuenan@pjlab.org.cn}

\author[label1]{Yingmei Wei\corref{cor1}}
\ead{weiyingmei@nudt.edu.cn}

\author[label1]{Jiangming Chen}
\ead{jiangming_chen@126.com}

\author[label1]{Yanming Guo}
\ead{guoyanming@nudt.edu.cn}
\cortext[cor1]{Corresponding authors.}

\affiliation[label1]{
    organization={Laboratory for Big Data and Decision, National University of Defense Technology},
    city={Changsha},
    state={Hunan},
    country={China}
}

\affiliation[label2]{
    organization={Shanghai AI Laboratory},
    city={Shanghai},
    state={Shanghai},
    country={China}
}



\begin{abstract}
Event cameras capture sparse brightness changes with high temporal resolution and high dynamic range, compensating for the deficiencies of the conventional RGB frames. 
However, previous multi-modal fusion techniques typically fail to handle the inherent heterogeneity between RGB frames and event streams, thus easily leading to noise amplification or redundant feature integration during cross-modal fusion. In this paper, we propose a Cross-Modal information inTeraction transFormer, coined as CMTFormer, which hierarchically integrates RGB and event information to achieve efficient and stable multimodal collaboration. Specifically, we design a shallow-to-deep information interaction scheme. In the shallow stage, we present the Shallow Alignment Module (SAM) to achieve an efficient fusion of RGB and event low-level features, which mitigates attribute disparities and prevents noisy information. In the middle stage, we devise the Cross-modal Enhancement Module (CEM) that utilizes texture and edge information to produce mutually reinforced middle-level features. In the deep stage, we present the Learnable Deep Fusion Module (LDFM) which performs high-level information aggregation through learnable weights, thus enabling the network to adaptively fuse RGB and event clues. A Spatial Prior Module is further designed to utilize global spatial information to enhance localization accuracy.
Extensive experiments are conducted on two prevalent event-based object detection benchmarks, \ie, DSEC-Detection and PKU-DAVIS-SOD. Our \algorithmname~consistently surpasses the detection counterparts in both uni-modal and multi-modal settings, strongly demonstrating the effectiveness of our paradigm. Codes will be available upon publication.
\end{abstract}




\end{frontmatter}



\section{Introduction}
Object detection plays a pivotal role in autonomous driving and embodied perception tasks~\cite{zou2023object,zhou2024yolo,sad,bev-survey,saratr-x,moe3d}. However, it often suffers from substantial performance degradation under adverse and challenging conditions, such as motion blur, low illumination, and dynamic lighting~\cite{hu2020learning}. In such scenarios, conventional frame-based cameras struggle to capture clear and informative visual cues due to their limited light sensitivity and inherent temporal latency. To overcome these constraints, event-based cameras~\cite{chakravarthi2024recent} have gained increasing attention as a complementary sensing modality~\cite{liu2025spiking,kang2025adaptive}, offering high temporal resolution, low latency, low power consumption, and intrinsic robustness to challenging conditions. How to effectively fuse the heterogeneous and asynchronous characteristics of these two modalities to achieve robust object detection remains a challenging and unresolved problem.

\begin{figure}[t!]
    \centering
    \includegraphics[width=1\linewidth]{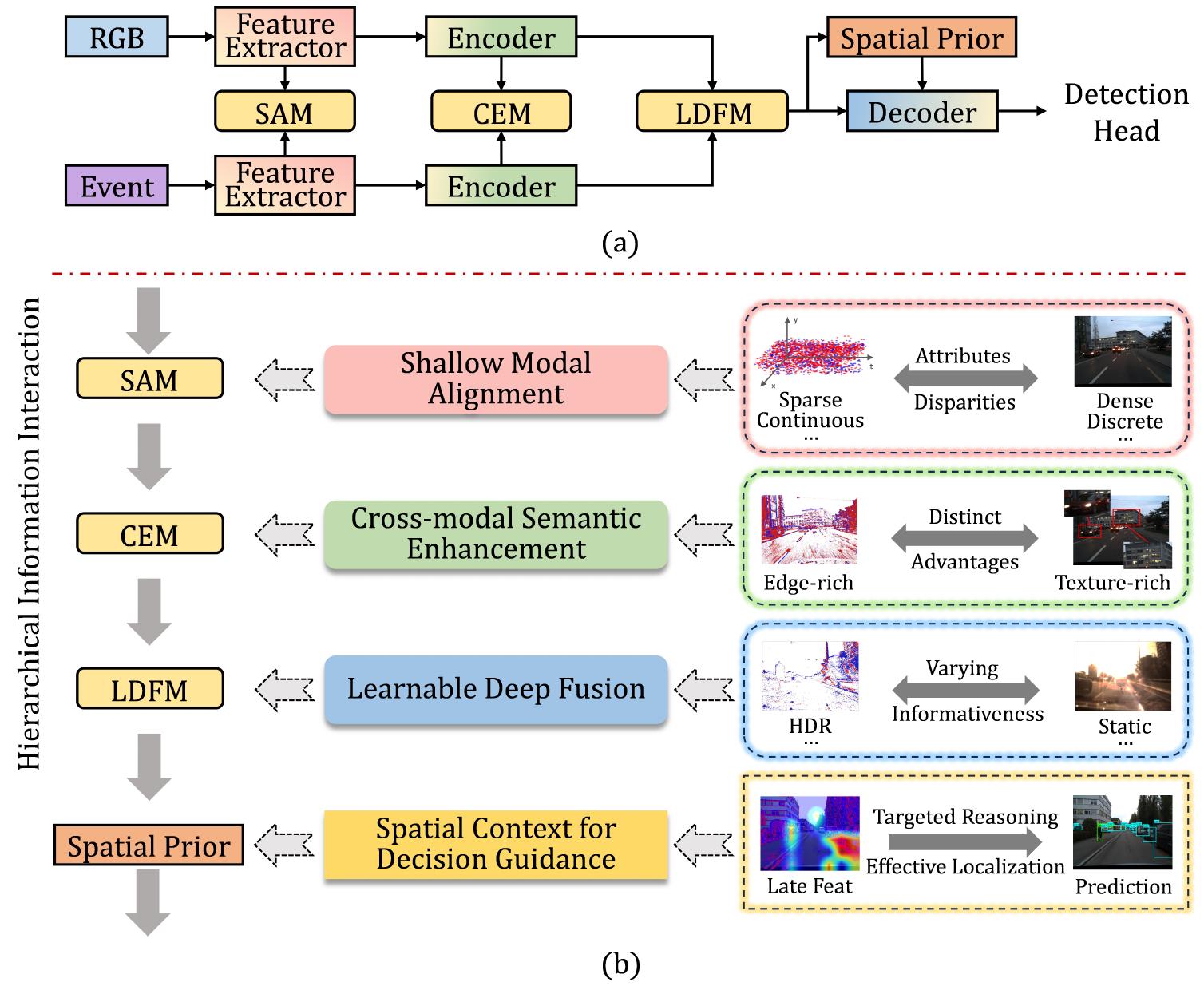}
    \caption{Schematic overview of our \algorithmname. (a) The architectural overview. We present the shallow-to-deep cross-modal fusion scheme.
    (b) Design motivations of our proposed modules. 
    }
    \label{fig:intro}
\end{figure}

Prior works on RGB–event fusion for object detection can be broadly categorized into four paradigms: early fusion~\cite{cao2021fusion, tomy2022fusing}, middle fusion~\cite{li2023sodformer, jiang2023nighttime}, late fusion~\cite{li2019event, jiang2019mixed} and hierarchical fusion~\cite{li2024hdi, cao2024embracing}.
Early fusion methods~\cite{cao2021fusion, tomy2022fusing} merge RGB frames and event streams after their respective backbone networks. 
However, they often suffer from semantic conflicts and noise coupling, since low-level features from the two modalities exhibit large modality gaps, which may lead to unstable or suboptimal representations.
The middle-fusion approaches~\cite{li2023sodformer, jiang2023nighttime} perform cross-modal interaction in the feature space. 
However, a single middle-stage fusion remains insufficient to fully exploit the complementary properties of RGB and event data, which limits the ability of the network to refine cross-modal alignment. 
Late fusion~\cite{li2019event, jiang2019mixed} integrates the two modalities only at the prediction stage. 
However, they often fail to exploit fine-grained cross-modal complementary cues, limiting performance improvements.
Hierarchical fusion methods~\cite{li2024hdi, cao2024embracing} apply multi-modal interaction at multiple stages of the pipeline. However, they typically reuse the same fusion module across stages, lacking stage-specific design and failing to account for the distinct roles and characteristics of different fusion levels.

These limitations reveal a fundamental challenge: the complementary information of RGB frames and event streams arises at different semantic levels, making single-stage fusion inherently insufficient for robust multi-modal detection. Moreover, when multi-modal fusion is performed across multiple stages, it requires stage-specific and purposely designed mechanisms. Uniform fusion operations fail to exploit the distinct roles and characteristics of each fusion level.
This motivates the need for a progressive fusion mechanism that conducts targeted modal integration at different stages to exploit the inter-modal characteristics and complementary strengths.

To this end, we propose \algorithmname, which marries transformer with hierarchical cross-modal information interaction to address modality imbalance and fully leverage cross-modal complementarity. The schematic overview is shown in Fig.~\ref{fig:intro}. 
Specifically, our hierarchical information fusion strategy consists of three key modules:
\textbf{Shallow Alignment Module (SAM)} aligns RGB and event modalities in the low-level feature space through lightweight interactions, mitigating attributes disparities while preventing noise caused by premature strong fusion.
\textbf{Cross-modal Enhancement Module (CEM)} leverages the distinct strengths of the two modalities—sharp edge dynamics from events and rich texture cues from RGB frames to realize bidirectional deep semantic complementarity. We introduce a Texture-Guided Enhancer and an Edge-Guided Enhancer. The former injects RGB texture information into event features, while the latter transfers event-driven edge structures into the RGB branch, producing mutually reinforced mid-level representations.
\textbf{Learnable Deep Fusion Module (LDFM)} performs high-level aggregation via learned attention weights, enabling the network to adaptively emphasize RGB or event cues depending on their reliability. 

We further propose a \textbf{Spatial-Prior Module}, which incorporates global spatial context to steer the decoding process, enabling the detector to perform informed and targeted reasoning. Meanwhile, it leverages temporal coherence across video frames to improve the localization accuracy. Extensive experiments on DSEC-Detection~\cite{gehrig2024low} and PKU-DAVIS-SOD~\cite{li2023sodformer} benchmarks demonstrate that our \algorithmname~significantly outperforms state-of-the-art detectors under both uni-modal and multi-modal conditions.

In summary, our contributions are listed below.
\begin{itemize}
    \item We propose a cross-modal information interaction transformer that hierarchically integrates RGB and event modalities from shallow to deep stages, enabling coarse-to-fine multi-modal fusion that preserves modality-specific strengths while enhancing fused representations.
    \item We introduce a Spatial-Prior Module that leverages global spatial context to guide the decoding process, improving query effectiveness by reducing redundant search and enabling more focused attention on target objects.
    \item We perform extensive experiments on DSEC-Detection~\cite{gehrig2024low} and PKU-DAVIS-SOD~\cite{li2023sodformer} benchmarks, where our \algorithmname~achieves state-of-the-art performance and demonstrates strong generalization under adverse conditions.
\end{itemize}

\section{Related Work}
\subsection{Event-based Object Detection}
Event-based object detection has gained increasing attention as an effective alternative to conventional frame-based detection in dynamic and high-speed scenarios, owing to the high temporal resolution, low latency, and high dynamic range of event cameras.
\cite{peng2023better} proposed the Hyper Histogram and Adaptive Event Conversion (AEC) modules, which preserve polarity and fine-grained temporal cues while improving detection accuracy and speed. 
RVT~\cite{gehrig2023recurrent} proposed a pioneering event-based detection backbone that significantly reduces inference latency while maintaining competitive accuracy. 
In addition to the above methods, some works have attempted to directly process raw event streams using Spiking Neural Networks (SNNs) or Graph Neural Networks (GNNs).
\cite{bodden2024spiking} presented Spiking CenterNet, which improves SNN detection performance via knowledge distillation. AEGNN~\cite{schaefer2022aegnn} proposed an asynchronous event-based GNN method that constructs a spatiotemporal graph from event streams for robust object detection in complex scenarios. 
Although these models are biologically inspired and energy-efficient, they often lag behind RGB-based methods in terms of detection accuracy and convergence speed.
Despite the progress in event-based detection, most current methods still struggle to capture semantic richness and maintain robustness in complex motion scenes, highlighting the need for more effective multimodal object detection methods that can jointly exploit the complementary strengths of cross-modalities.

\subsection{RGB-Event Fusion for Object Detection}
Recent advances in multimodal object detection have explored fusing RGB and event data~\cite{chitta2022transfuser, tulyakov2021time, duan2021guided, zhang2021object, gehrig2021combining, zuo2022devo, gao2022vector} to leverage their complementary properties—dense semantic textures from RGB frames and fine-grained motion cues from event streams. 
Early studies primarily relied on post-detection fusion, where bounding boxes or confidence maps from each modality were combined at the decision level via weighting or voting strategies.
\cite{li2019event} integrates frame-based and event-based detections at the decision level by combining bounding boxes and confidence scores, aiming to improve robustness in challenging driving scenarios. 
While these approaches offer modularity and implementation flexibility, they lack deep interaction between modalities during the feature learning process. As a result, they fall short in capturing joint spatial-temporal correlations and fail to fully exploit the complementary advantages of event streams and RGB images.
To capture cross-modal complementary cues, some works advance the fusion stage into the intermediate stage, enabling feature interaction through attention mechanisms or explicit alignment strategies.
\cite{cao2024embracing} proposed a Hierarchical Feature Refinement Network that processes RGB and event modalities through separate backbones and applies fusion modules at multiple feature levels to enhance representation under challenging conditions. \cite{cao2021fusion} employed a feature attention gate for vehicle detection using event cameras, achieving better robustness in complex driving scenes. Other works leverage attention mechanisms to guide modality interaction. For example, \cite{liu2021attention} designed an Attention Fusion Network (AFN) for event-based vehicle detection, where cross-modality attention selectively integrates event features into RGB streams. 
SODFormer~\cite{li2023sodformer} integrates asynchronous RGB and event streams using a temporal transformer and spatial fusion modules to model cross-frame consistency and modality alignment. Additionally, spiking neural networks have been used to build energy-efficient fusion frameworks, as in ESNN-Fusion~\cite{fan2025efficient}, which combines RGB and event information via spike-based attention units.
Despite these efforts, many existing models perform fusion either at a fixed semantic stage or treat both modalities symmetrically, without fully exploring their cross-modal interactions.

\section{Method}
In this section, we first present an overview of our \algorithmname~in Sec.~\ref{subsec:networkoverview}. Then, we introduce the basic knowledge in Sec.~\ref{subsec:eventrepresentation}, including the working principles of event cameras and common event representations. Afterward, Sec.~\ref{subsec:trilevelfusion} illustrates our hierarchical cross-modal information interaction strategy. Finally, Sec.~\ref{subsec:perceptionguided} describes our spatial-prior module.

\subsection{Network Overview}
\label{subsec:networkoverview}

\begin{figure*}[t!]
    \centering
    \includegraphics[width=\textwidth]{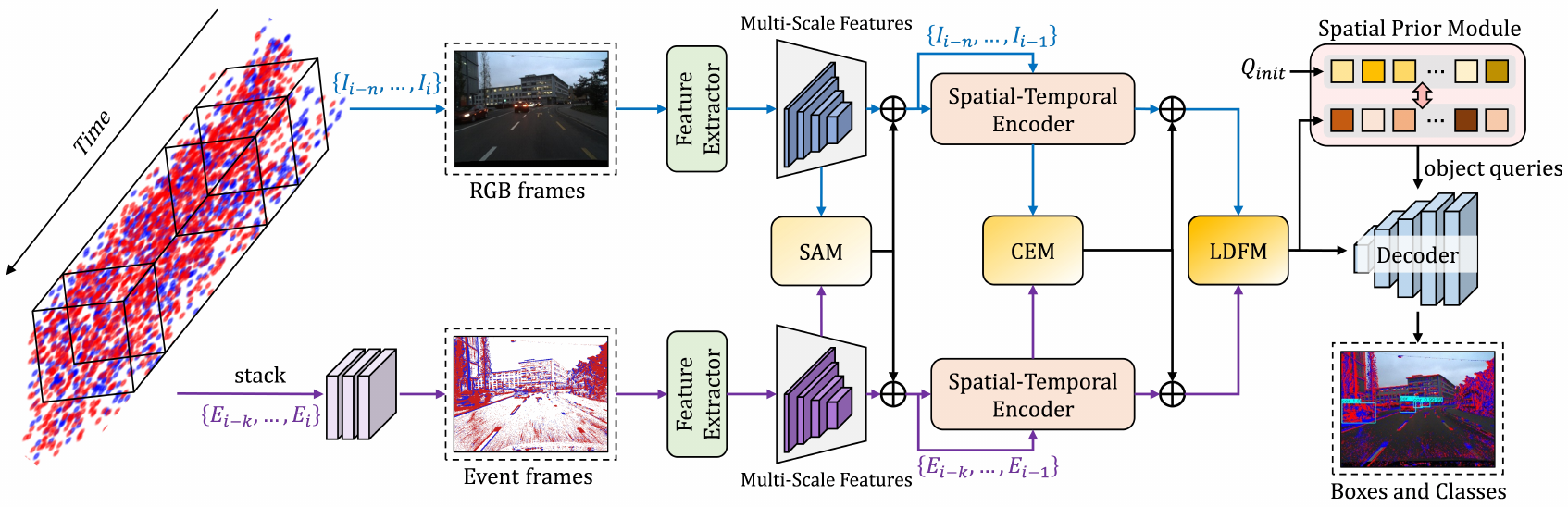}
    \caption{Framework overview of our \algorithmname. Our model takes stacked event frames and RGB frames as inputs and extracts multi-scale features through a dual-stream feature extractor. A progressive fusion strategy is then applied across the network: the Shallow Alignment Module (SAM) performs coarse low-level interaction on the features extracted from both modalities, the Cross-modal Enhancement Module (CEM) conducts bidirectional semantic enhancement within the encoder, and the Learnable Deep Fusion Module (LDFM) fuses high-level representations adaptively based on the learned attention weights before the decoder. The fused features are further leveraged by the Spatial-Prior Module, which guides the decoding process. The final detection results are produced by the detection head. 
    }
    \label{fig:overall_architecture}
\end{figure*}


The pipeline of our \algorithmname~is illustrated in Fig.~\ref{fig:overall_architecture}. Specifically, the continuous event stream is first accumulated into event frames using temporal binning, enabling compatibility with standard convolutional. The event frames, together with the corresponding RGB frames, are fed into a dual-stream feature extractor to obtain modality-specific multi-scale features. The extracted features are subsequently fused via a progressive strategy. At the early stage, the Shallow Alignment Module (SAM) operates on the outputs of the feature extractor to align low-level structures between RGB and events. The fused feature map is then passed into spatial–temporal encoders, which model both intra-frame spatial context and inter-frame temporal dependencies. Within the encoder, the Cross-modal Enhancement Module (CEM) implements mid-level semantic interaction, leveraging the distinct advantages of the two modalities. Before the decoder, the Learnable Deep Fusion Module (LDFM) uses learned attention weights to adaptively select and combine modality-specific high-level cues, producing a unified representation that emphasizes the most informative modality at each spatial location.
Subsequently, the Spatial-Prior Module leverages the late fused feature to provide global guidance for downstream decoding. Finally, the detection head produces the final object categories and bounding boxes.

\subsection{Event Representation}
\label{subsec:eventrepresentation}

Event cameras output a continuous stream of asynchronous brightness changes,
\begin{equation}
\mathcal{E}=\{(x_i, y_i, t_i, p_i)\mid i=1,\dots,N\},
\end{equation}
where $(x_i,y_i)$ denotes the pixel location, $t_i$ is the timestamp, and $p_i\in\{-1,+1\}$ is the event polarity.
To convert this asynchronous stream into a format compatible with 2D neural networks, 
we divide the events within a temporal window $[t,\, t+\Delta t]$ into $B$ uniformly spaced bins.
Each event is assigned to a bin according to
\begin{equation}
b_i = \left\lfloor 
B \cdot \frac{t_i - t}{\Delta t}
\right\rfloor , \qquad b_i \in \{0,\dots,B-1\},
\end{equation}
and events in the same bin are accumulated spatially to form an image-like slice.
For each temporal bin $b\in\{0,\dots,B-1\}$, we accumulate all events assigned to this bin into a 2D spatial map
\begin{equation}
E(b,y,x)
=\sum_{i:\, b_i=b} \delta(x-x_i)\,\delta(y-y_i)\,p_i,
\end{equation}
where $\delta(\cdot)$ is the Kronecker delta function~\cite{graham1994concrete}.
Stacking all bin-wise maps along the channel dimension produces the event tensor.
\begin{equation}
\mathbf{E} = \big[\,E(0),\,E(1),\,\dots,\,E(B-1)\,\big]
\;\in\;\mathbb{R}^{B\times H\times W},
\end{equation}
where $H$ and $W$ denote the spatial resolution of the event frame.
This representation retains fine-grained temporal structure while enabling efficient processing using standard CNN or Transformer backbones.

Compared with other event representations such as voxel grids~\cite{zhu2019unsupervised} or sigmoid representation\cite{chen2018pseudo}, temporally binned event frames achieve a favorable balance between accuracy and efficiency. In our framework, the resulting event frames serve as the input to the event branch of the dual-stream feature extractor and participate in the subsequent progressive multi-stage fusion with RGB features.

\subsection{Hierarchical Cross-Modal Information Interaction}
\label{subsec:trilevelfusion}
We propose a progressive cross-modal fusion strategy to fully leverage the complementary properties of RGB and events at different levels. 
It consists of three dedicated modules, \ie, the Shallow Alignment Module (SAM), the Cross-modal Enhancement Module (CEM), and the Learnable Deep Fusion Module (LDFM), which operate sequentially to facilitate progressive cross-modal integration.

\begin{figure}[htbp!]
    \centering
    \includegraphics[width=1\linewidth]{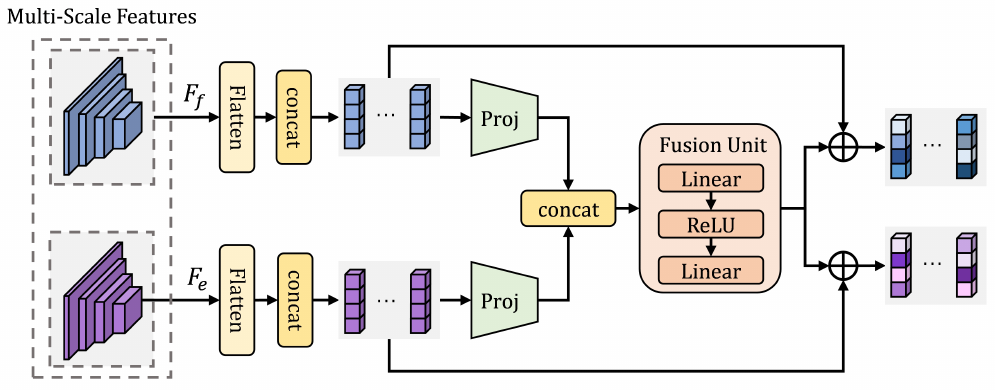}
    \caption{Illustration of the Shallow Alignment Module (SAM), a lightweight design for coarse modal alignment that prevents noise amplification.}
    \label{fig:early_fusion}
\end{figure}

\paragraph{\textbf{Shallow Alignment Module (SAM)}}
Although RGB frames and event streams exhibit strong complementarity, directly fusing them at the early feature stage is challenging due to substantial modality discrepancies in spatial density, temporal characteristics, and noise distribution. 
Aggressive early fusion often disrupts each modality’s intrinsic representation, leading to feature contamination or semantic misalignment in deeper layers. 
To address this, SAM is designed as a lightweight alignment module that performs only coarse and structurally focused interaction. 
By aligning multi-scale low-level cues without enforcing strong semantic mixing, SAM reduces cross-modal inconsistencies while preserving modality-specific information, thereby providing a stable and well-aligned foundation for the subsequent semantic interaction performed in the middle fusion stage. The architecture of SAM is illustrated in Fig.~\ref{fig:early_fusion}.

The RGB frames and event frames are first processed by a dual-stream feature extractor, yielding multi-scale features 
$\{\mathcal{F}^{l}_{f}\}_{l=1}^{L}$ and $\{\mathcal{F}^{l}_{e}\}_{l=1}^{L}$,
where each level $l$ produces a feature map $\mathcal{F}^{l} \in \mathbb{R}^{H_l \times W_l \times C_l}$.
To enable unified low-level interaction, all features are flattened and concatenated across scales and subsequently projected into a shared embedding space. 
The aligned features are then fused through a shallow interaction unit, formulated as:
\begin{equation}
\tilde{\mathcal{F}} = 
\varphi \!\left(
    \left[
        \mathcal{W}_{f}\!\left(\Theta_{l}\!\left(\Gamma(\mathcal{F}^{l}_{f})\right)\right),\;
        \mathcal{W}_{e}\!\left(\Theta_{l}\!\left(\Gamma(\mathcal{F}^{l}_{e})\right)\right)
    \right]
\right),
\end{equation}
where 
$\Gamma(\cdot)$ represents the flattening operation,
$\Theta_{l}(\cdot)$ denotes multi-scale concatenation across all levels,
$\mathcal{W}_{f}$ and $\mathcal{W}_{e}$ denote linear projection layers for RGB and event features and
$\varphi(\cdot)$ is a lightweight two-layer MLP.

Finally, SAM applies residual modulation to refine both modalities:
\begin{equation}
\hat{\mathcal{F}}_{f} = \mathcal{F}_{f} + \alpha_{f}\,\tilde{\mathcal{F}},
\qquad
\hat{\mathcal{F}}_{e} = \mathcal{F}_{e} + \alpha_{e}\,\tilde{\mathcal{F}},
\end{equation}
where $\alpha_{f}$ and $\alpha_{e}$ are learnable fusion coefficients that regulate the contribution of the fused representation to each modality.

\begin{figure}[htbp!]
    \centering
    \includegraphics[width=\linewidth]{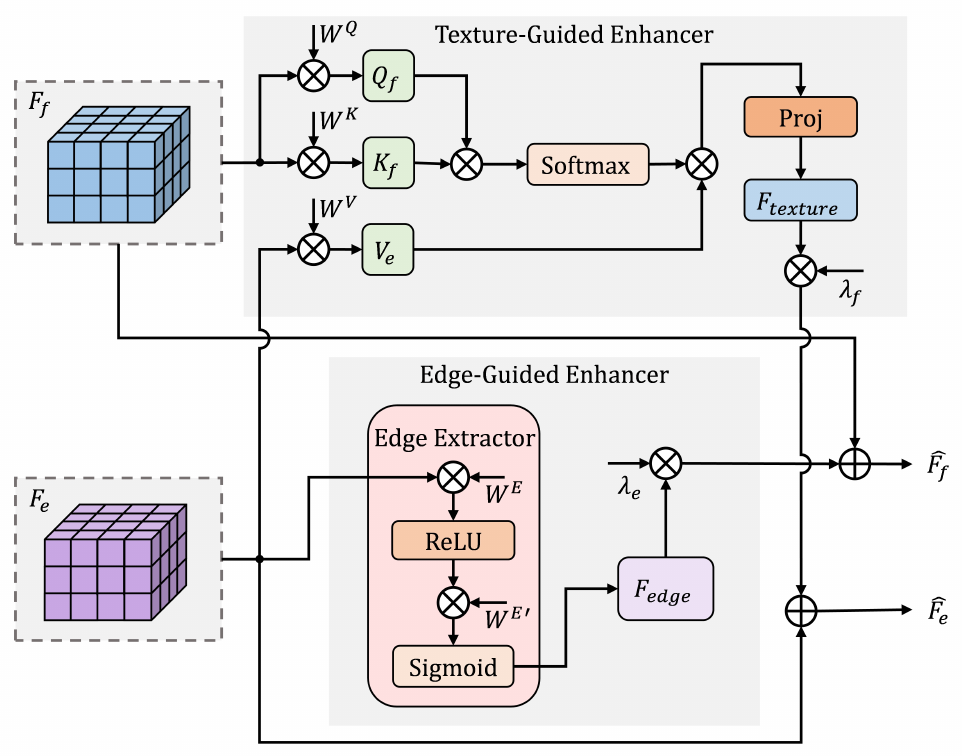}
    \caption{Illustration of the Cross-modal Enhancement Module (CEM). CEM performs bidirectional enhancement between RGB and event features, using RGB texture cues to enhance event representations and event edge cues to enhance RGB features, thereby strengthening semantic complementarity across modalities.}
    \label{fig:middle_fusion}
\end{figure}

\paragraph{\textbf{Cross-modal Enhancement Module (CEM)}}
After the coarse modal alignment provided by SAM, CEM performs mid-level semantic enhancement between RGB and event features. Its primary objective is to exploit the complementary strengths of the two modalities: RGB frames supply texture and appearance cues, whereas event streams highlight edge structures and motion dynamics. To effectively leverage this complementarity, CEM is composed of two submodules, \ie, Texture-Guided Enhancer and Edge-Guided Enhancer. The former enriches event representations with RGB texture information, while the latter injects event-driven edge cues into RGB features. Through this bidirectional interaction, CEM yields mutually enhanced and semantically aligned feature representations. The overall architecture of CEM is illustrated in Fig.~\ref{fig:middle_fusion}.

To enrich event features with RGB texture cues, we compute cross-attention between RGB and event features to derive a texture-guided representation $\mathcal{F}_{\text{texture}}$, which selectively transfers texture and appearance information from the RGB branch to the event branch. This representation is then incorporated into the event features via residual modulation, yielding the enhanced event feature $\hat{\mathcal{F}}_e$.
Let $\mathcal{F}_f$ and $\mathcal{F}_e$ denote the mid-level RGB and event features from the transformer encoder.
The texture-guided enhancemer is formulated as:
\begin{align}
&\mathcal{F}_{texture} = \operatorname{softmax} \left(\frac{Q_f K_f^\top}{\sqrt{d}}\right) V_e, \\
&\hat{\mathcal{F}}_e = \mathcal{F}_e + \lambda_f \, \mathcal{F}_{texture},
\end{align}
\begin{equation}
Q_f = \mathcal{F}_f W^{Q}, \quad 
K_f = \mathcal{F}_f W^{K}, \quad 
V_e = \mathcal{F}_e W^{V},
\end{equation}
where $W^{Q}$, $W^{K}$, and $W^{V}$ are learnable projection matrices that generate the query, key, and value embeddings for cross-modal attention, $d$ denotes the dimensionality of the query/key vectors and $\lambda_f$ is a learnable modulation coefficient.

Event streams provide fine-grained motion dynamics and sharp edge structures. To inject these discriminative cues into the RGB branch, we employ an edge extractor that generates an event-driven edge feature $\mathcal{F}_{\text{edge}}$ from the event representation. This edge feature is subsequently incorporated into the RGB branch through residual modulation, resulting in an enhanced RGB feature $\hat{\mathcal{F}}_f$ with improved structural sensitivity.
The edge-guided enhancer for RGB representations is defined as:
\begin{align}
&\mathcal{F}_{edge} =
\operatorname{sigmoid}\left(
\mathcal{W}^{E'} \big( \phi \big( \mathcal{W}^{E}(\mathcal{F}_e) \big) \big)
\right), \\
&\hat{\mathcal{F}}_f =
\mathcal{F}_f + \lambda_e \, \mathcal{F}_{edge},
\end{align}
where $\mathcal{W}^{E}$ and $\mathcal{W}^{E'}$ are learnable linear projections,
$\phi(\cdot)$ denotes the ReLU activation, and $\lambda_e$ is a learnable modulation coefficient.

\begin{figure}[htbp!]
    \centering
    \includegraphics[width=0.7\linewidth]{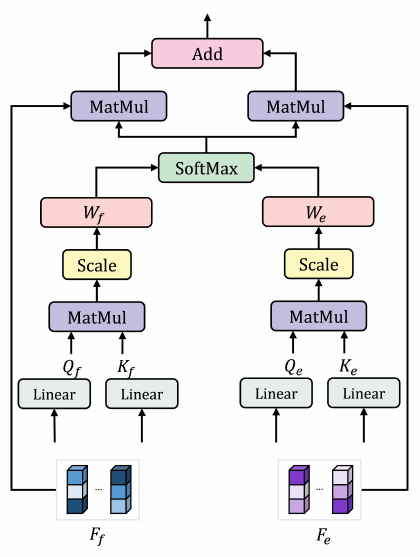}
    \caption{The architecture of the Learnable Deep Fusion Module (LDFM). LDFM adaptively integrates high-level RGB and event features through learned attention weights, enabling the network to emphasize the modality most informative for each spatial location under different conditions.}
    \label{fig:late_fusion}
\end{figure}

\paragraph{\textbf{Learnable Deep Fusion Module (LDFM)}}
After the bidirectional semantic enhancement performed in CEM, the Learnable Deep Fusion Module (LDFM) further refines RGB and events integration at a scene level.
Unlike early or middle fusion, which focuses on structural alignment or semantic enhancement, LDFM aims to adaptively determine which modality is more informative at each spatial location. This is essential because the reliability of each modality fluctuates across scenarios: RGB may degrade under motion blur or low-light, whereas events may become noisy in cluttered or stationary regions.
To address this, LDFM learns modality-specific attention weights to adaptively integrate RGB and event features. The architecture of LDFM is illustrated in Fig.~\ref{fig:late_fusion} and formulated as follows.
\begin{align}
    {Attn}_{f} &= \mathrm{Softmax} \left( \frac{Q_{f}K_{f}^\top}{\sqrt{d}} \right), \\      
    {Attn}_{e} &= \mathrm{Softmax} \left( \frac{Q_{e}K_{e}^\top}{\sqrt{d}} \right), \\
    F_{fused} &= \eta \, {Attn}_{f}V_{f} 
    + (1-\eta) \, {Attn}_{e}V_{e},
\end{align}
where \(Q_{f}\), \(K_{f}\), \(V_{f}\) are the query, key, and value projections of RGB features, and \(Q_{e}\), \(K_{e}\), \(V_{e}\) are those of event features. $d$ denotes the dimensionality
of the query/key vectors. The scalar \(\eta\) is a learnable parameter that balances the contributions from each modality.

This adaptive fusion mechanism allows the network to dynamically balance the contributions of RGB and event features based on their informativeness. The fused representation \(F_{fused} \) is then fed into the decoder for final object prediction.

\subsection{Spatial-Prior Module}
\label{subsec:perceptionguided}
The Spatial-Prior Module is designed to enhance localization accuracy and reduce redundant search during object decoding by injecting global spatial context into the queries. 
Conventional DETR-style decoders rely on a set of learnable queries that are randomly initialized and require multiple decoding layers to gradually discover objects. However, this blind and iterative search often leads to slow convergence and unstable localization in challenging scenarios such as fast motion or low illumination. 

\begin{figure}[htbp!]
    \centering
    \includegraphics[width=1\linewidth]{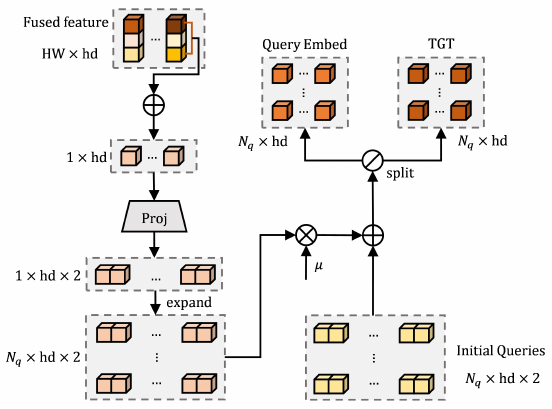}
    \caption{Overview of the Spatial-Prior Module, which uses the global context to modulate the initial queries, enabling targeted decoding that enhances localization effectiveness.}
    \label{fig:pgst}
\end{figure}

To address these limitations, we introduce a Spatial-Prior Module that adaptively modulates the decoder queries using global spatial context extracted from both RGB and event features which include both spatial and temporal cues. The architecture is illustrated in Fig.~\ref{fig:pgst}. Specifically, after LDFM produces the fused feature, we perform global average pooling followed by a projection layer to obtain a compact global descriptor that encodes holistic scene semantics and motion context. This descriptor is used to modulate the initial query embeddings, producing a set of perceptive queries that are dynamically conditioned on the input.

Specifically, given the fused cross-modal feature map $\mathcal{F}_{fused}$, we first perform global average pooling to extract a global-contextual vector. This vector is then projected into the query embedding space. To align with the number of decoder queries, the projected vector is broadcast across all $N_q$ query slots, denoted as $q_{global}$. The global-contextual queries are then combined with the random initial learnable queries $q_{init}$ through an adaptive weighting mechanism controlled by a coefficient $\mu$. The perceptive query is denoted as $q_{percep}$. The procedure is formulated as:
\begin{align}
    q_{global} &= R\!\left(\psi\!\left(\Omega(\mathcal{F}_{fused})\right),\, N_q\right), \\
    q_{percep} &= \mu\, q_{global} + q_{init} ,
\end{align}
where $\Omega(\cdot)$ denotes global average pooling, $\psi(\cdot)$ is a linear projection layer, and $R(\cdot, N_q)$ replicates the projected vector to match the number of queries $N_q$. 
Finally, the embeddings $q_{percep}$ are split into the query and target vectors used by the decoder.

\section{Experiments}
Sec.~\ref{section:A} describes the experimental setting and implementation details. Sec.~\ref{section:B} reports the quantitative performance of our approach under diverse conditions, along with comparisons to state-of-the-art methods. We perform comprehensive ablation studies in Sec.~\ref{section:C} to analyze the effects of each module and key parameter choices. Finally, Sec.~\ref{section:D} presents the scalability and interpretability of our framework.

\subsection{Experimental Settings}
\label{section:A}
\paragraph{Dataset}  
We conduct experiments on the DSEC-Detection dataset~\cite{gehrig2024low} and PKU-DAVIS-SOD dataset~\cite{li2023sodformer}. DSEC-Detection dataset~\cite{gehrig2024low} provides paired RGB frames and events at a resolution of 640×480, along with 390 K boxes at 20 Hz for object detection in driving scenarios. PKU-DAVIS-SOD~\cite{li2023sodformer} is a large-scale event-based object detection benchmark specifically designed for high-speed and low-light scenarios. It provides asynchronous events, RGB frames, and manual labels at 25 Hz. The resolution is 346×260 and the total number of boxes is 1080.1 K.

\paragraph{Implementation Details}
We adopt event images as the input representation for the event modality, following\cite{maqueda2018event}, and utilize ResNet-50\cite{he2016deep} as the feature extractor for both RGB and event branches due to its favorable balance between accuracy and computational efficiency. During training, the input image height is randomly scaled to a number from [256, 576] and the interval is 32, while the width is adjusted proportionally to maintain the aspect ratio. For evaluation, all images are resized to 352 in height and 468 in width to maintain the aspect ratio. We avoid additional augmentations such as random cropping or horizontal flipping, as the accuracy of reference point alignment is critical to the performance of the temporal attention module. Following the training strategy in Deformable DETR\cite{zhu2020deformable}, we adopt a Hungarian matching loss with weight coefficients of 2, 5, and 2 for the classification loss, L$_{1}$ loss, and GIoU loss, respectively.
The model is optimized using AdamW\cite{kinga2015method} with an initial learning rate of \(2\times 10^{-4} \), a weight decay of \(10^{-4}\), and a learning rate drop by a factor of $0.1$ after the 20th epoch. All experiments are conducted using NVIDIA RTX 4090 GPUs with a batch size of 8.

\paragraph{Evaluation Metrics} 
We adopt the mean Average Precision (mAP)\cite{lin2014microsoft} as metrics. Specifically, in the effective test, we use average precision across multiple Intersection-over-Union (IoU) thresholds (\ie, AP, AP$_{0.5}$, AP$_{0.75}$) and multi-scale average precision (\ie, AP$_{S}$, AP$_{M}$, AP$_{L}$) for evaluation. In other tests, AP$_{0.5}$ is uniformly used as the reported detection performance metric. For performance evaluation in various scenarios, we calculate the AP for each category and derive the mAP value for three categories.

\subsection{Effective Test}
\label{section:B}

\paragraph{Performance Evaluation in Various Scenarios}
Table~\ref{tab:pkudavis} presents a comprehensive quantitative evaluation of detection performance on the PKU-DAVIS-SOD dataset~\cite{li2023sodformer} across normal, motion blur, and low-light scenarios. The results demonstrate that multimodal fusion of RGB frames and events consistently surpasses single-modality approaches in terms of mAP and other metrics for all object categories. Notably, under challenging conditions such as motion blur and low-light, the fusion strategy yields substantial improvements, underscoring the complementary strengths of RGB and event modalities. These findings confirm that the proposed method achieves superior robustness and accuracy in complex real-world environments.

\begin{table*}[t!]
\centering
\caption{
Performance evaluation on the PKU-DAVIS-SOD dataset~\cite{li2023sodformer} under different scenarios.
The table reports detection results for three modality settings: Events only, RGB only, and RGB + Events. The single-modality baselines rely only on RGB or event inputs without using any fusion strategy or spatial-prior guided design, whereas the multimodal setting corresponds to our \algorithmname, which integrates both modalities through a spatial-prior guided progressive fusion mechanism. All metrics are evaluated across normal, motion-blur, and low-light conditions.}
\label{tab:pkudavis}
\resizebox{\textwidth}{!}{
\begin{tabular}{llccccccccc}
\toprule
\multirow{2}{*}{Scenario} & \multirow{2}{*}{Modality} 
& \multicolumn{3}{c}{AP$_{50}$} 
& \multirow{2}{*}{mAP} & \multirow{2}{*}{mAP$_{50}$} 
& \multirow{2}{*}{mAP$_{75}$} & \multirow{2}{*}{mAP$_S$} 
& \multirow{2}{*}{mAP$_M$} & \multirow{2}{*}{mAP$_L$} \\
\cmidrule(lr){3-5}
& & Car & Pedestrian & Two-wheeler & & & & & & \\
\midrule

\multirow{3}{*}{\cellcolor{white}Normal} 
& Events           & 0.440 & 0.220 & 0.451 & 0.147 & \textbf{0.371} & 0.090 & 0.072 & 0.268 & 0.526 \\
& RGB              & 0.747 & 0.365 & 0.558 & 0.228 & \textbf{0.557} & 0.138 & 0.166 & 0.336 & 0.539 \\
\rowcolor{blue!10}
\cellcolor{white}& RGB + Events     & 0.758 & 0.403 & 0.600 & 0.256 & \textbf{0.587} & 0.184 & 0.188 & 0.377 & 0.566 \\
\midrule

\multirow{3}{*}{\cellcolor{white}Motion blur} 
& Events           & 0.327 & 0.159 & 0.380 & 0.113 & \textbf{0.289} & 0.064 & 0.051 & 0.181 & 0.255 \\
& RGB              & 0.561 & 0.303 & 0.394 & 0.163 & \textbf{0.419} & 0.096 & 0.100 & 0.201 & 0.365 \\
\rowcolor{blue!10}
\cellcolor{white}& RGB + Events     & 0.576 & 0.285 & 0.452 & 0.185 & \textbf{0.437} & 0.131 & 0.114 & 0.235 & 0.369 \\
\midrule

\multirow{3}{*}{\cellcolor{white}Low-light} 
& Events           & 0.524 & 0.0002 & 0.294 & 0.093 & \textbf{0.273} & 0.039 & 0.075 & 0.183 & 0.286 \\
& RGB              & 0.570 & 0.128 & 0.357 & 0.114 & \textbf{0.351} & 0.048 & 0.082 & 0.198 & 0.344 \\
\rowcolor{blue!10}
\cellcolor{white}& RGB + Events     & 0.580 & 0.162 & 0.511 & 0.146 & \textbf{0.418} & 0.070 & 0.112 & 0.240 & 0.432 \\
\midrule

\multirow{3}{*}{\cellcolor{white}All} 
& Events           & 0.424 & 0.188 & 0.390 & 0.128 & \textbf{0.334} & 0.071 & 0.065 & 0.210 & 0.348 \\
& RGB              & 0.700 & 0.316 & 0.452 & 0.195 & \textbf{0.489} & 0.116 & 0.142 & 0.264 & 0.417 \\
\rowcolor{blue!10}
\cellcolor{white}& RGB + Events     & 0.711 & 0.340 & 0.525 & 0.223 & \textbf{0.525} & 0.155 & 0.164 & 0.313 & 0.450 \\
\bottomrule
\end{tabular}
}
\end{table*}

\begin{figure*}[t]
    \centering
    \includegraphics[width=\textwidth]{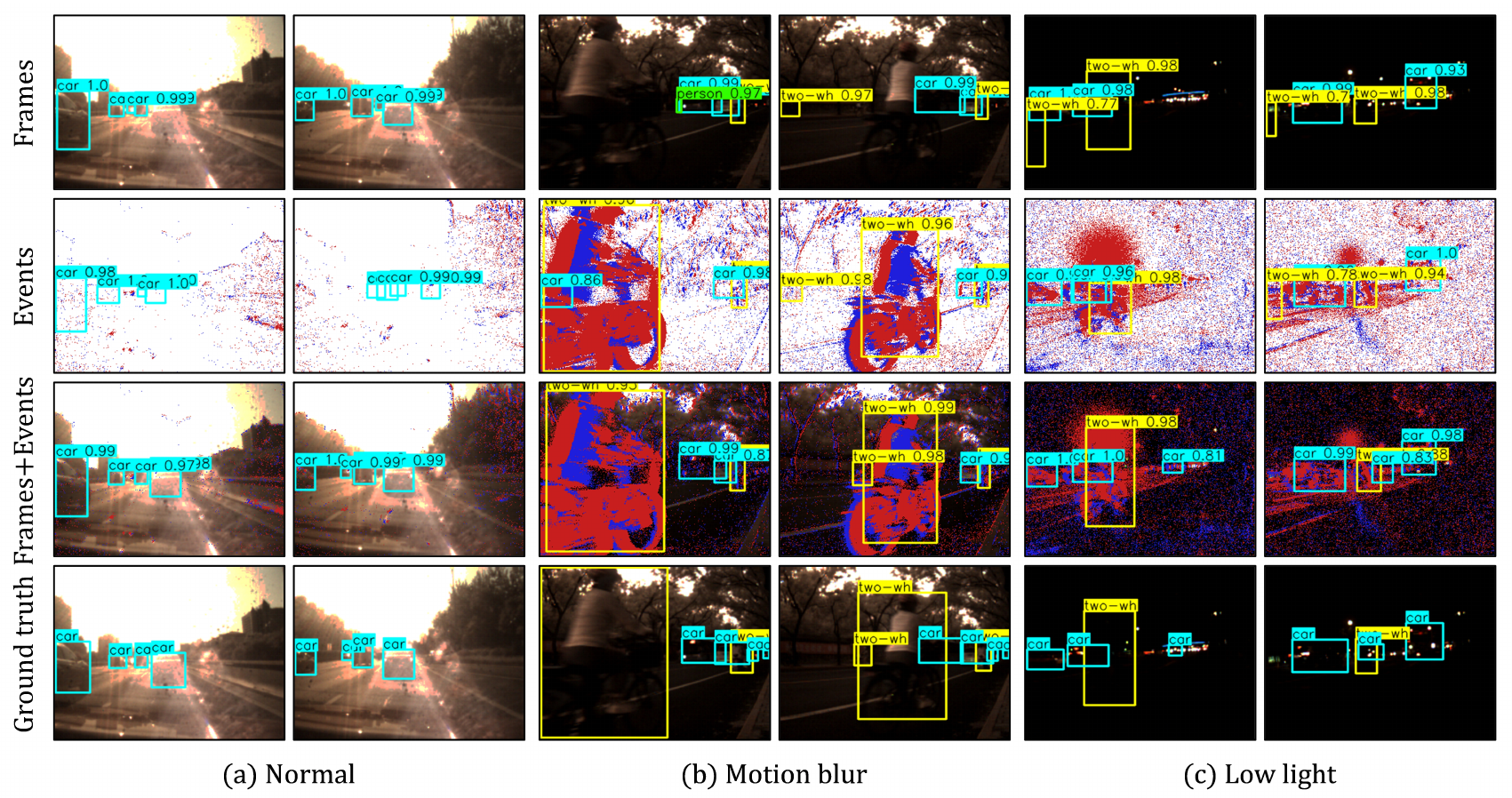}
    \caption{Visual predictions of \algorithmname~under different scenarios. The three columns from left to right display normal, high-speed motion blur, and low-light scenes, respectively. Our \algorithmname~achieves robust detection across all scenarios.}
    \label{fig:various_scene}
\end{figure*}

Fig.~\ref{fig:various_scene} provides qualitative visualizations of object detection results of our \algorithmname~under representative scenarios, including overexposure, high-speed motion blur and low-light conditions. Evidently, our approach maintains reliable detection performance by leveraging event-based cues to compensate for degraded RGB frames, enabling robust object detection even in adverse scenarios.

\paragraph{Comparison with State-of-the-Art Methods}
In this section, we provide a comprehensive comparison of our \algorithmname~with state-of-the-art methods on both DSEC-Detection~\cite{gehrig2024low} and PKU-DAVIS-SOD~\cite{li2023sodformer} datasets. We include recent leading approaches for event-based object detection, RGB-based detection, and multimodal fusion-based detection. For single-modality baselines, we utilize our spatial-prior guided \algorithmname, but exclude all components related to the progressive cross-modal fusion strategy.

\subparagraph{Evaluation on DSEC-Detection Dataset.}
To rigorously assess the effectiveness of the proposed \algorithmname, we conduct comprehensive comparisons against twelve state-of-the-art approaches and our baselines, as summarized in Table~\ref{tab:comparison1}. Notably, our method consistently outperforms existing methods across diverse conditions, demonstrating its advantages on three different modalities: RGB-only, event-only, and multimodal fusion.

Among eight competitive RGB-based detectors, including two-stage frameworks (Faster R-CNN~\cite{ren2016faster}), one-stage designs (RetinaNet~\cite{lin2017focal}, CenterNet~\cite{zhou2019objects}, YOLOv5~\cite{zhu2021tph}, YOLOv7~\cite{wang2023yolov7}, ConvNeXt~\cite{liu2022convnet}), and transformer-based approaches (Deformable DETR~\cite{zhu2020deformable}, Swin Transformer~\cite{liu2021swin}), our RGB baseline obtains the highest performance, reaching $0.494$ mAP$_{50}$. This demonstrates that our spatial-prior module significantly strengthen the detector, surpassing even large-capacity backbones such as Swin-T and Hourglass-104.

Event-driven detectors generally underperform RGB-based approaches due to sparse textures and high noise levels in outdoor scenes. Despite this, our event baseline achieves $0.291$ mAP$_{50}$, outperforming EMS-YOLO~\cite{su2023deep} by $8.9$\%, Deformable DETR~\cite{zhu2020deformable} by $10.4$\%, and Swin-based YOLO~\cite{liu2021swin} by $2.5$\%. These improvements clearly demonstrate the advantages of spatial-prior module, even under event-only configurations.

The multi-modal setting further reveals the full capability of our \algorithmname. Classical fusion strategies (\eg, FPN-fusion~\cite{tomy2022fusing} and SFNet~\cite{liu2024enhancing}) provide only moderate gains over unimodal detectors, while SODFormer~\cite{li2023sodformer} achieves stronger performance. In contrast, our \algorithmname~delivers the best results across all metrics, reaching $0.506$ mAP$_{50}$. Compared to SODFormer~\cite{li2023sodformer}, this corresponds to consistent gains of $2.8$\%, underscoring the effectiveness of our \algorithmname, which enables more robust cross-modal alignment and feature aggregation in dynamic and challenging conditions. Note that we do not compare our algorithm with more recent detectors, such as CAFR~\cite{cafr}, since they merely report performance on three classes. 

\begin{table*}[t]
\centering
\caption{Comparison with state-of-the-art methods and our \algorithmname~on the DSEC-Detection dataset \cite{gehrig2024low}. Methods annotated with * indicate the use of a spatial-prior module.}
\label{tab:comparison1}
\resizebox{\textwidth}{!}{
\begin{tabular}{llllccll}
\toprule
Modality & Method & Backbone & Head & Temporal & Spatial-Prior & \#Params & mAP$_{50}$ \\ 
\midrule
\multirow{9}{*}{\cellcolor{white}RGB} 
& Faster R-CNN \cite{ren2016faster} & ResNet50+FPN & Faster R-CNN & No & No & 41.9M & 0.354 \\
& RetinaNet \cite{lin2017focal} & ResNeXt-101-FPN & RetinalHead & No & No & 53.9M & 0.305 \\
& CenterNet \cite{zhou2019objects} & Hourglass-104 & CenterNet & No & No & 191.4M & 0.351 \\
& YOLOv5 \cite{zhu2021tph} & CSPDarknet53 & YOLOv5 & No & No & 76.8M & 0.332 \\
& YOLOv7 \cite{wang2023yolov7} & E-ELAN & YOLOv7 & No & No & 151.7M & 0.315 \\
& ConvNeXt \cite{liu2022convnet} & ConvNeXt-T & Mask R-CNN & No & No & 45.6M & 0.462 \\
& Deformable DETR \cite{zhu2020deformable} & DETR & DETR & No & No & 39.4M & 0.413 \\
& Swin Transformer \cite{liu2021swin} & Swin-T & Mask R-CNN & No & No & 44.8M & 0.487 \\
\rowcolor{blue!10}
\cellcolor{white}& Our baseline* & Deformable DETR & Deformable DETR & Yes & Yes & 46.4M & \textbf{0.494} \\
\midrule
\multirow{4}{*}{\cellcolor{white}Event} 
& EMS-YOLO \cite{su2023deep} & EMS-ResNet10 & YOLOv3 & No & No & 6.2M & 0.202 \\
& Deformable DETR \cite{zhu2020deformable} & DETR & DETR & No & No & 39.4M & 0.187 \\
& Swin Transformer \cite{liu2021swin} & Swin-T & YOLOv3 & No & No & 44.8M & 0.266 \\
\rowcolor{blue!10}
\cellcolor{white}& Our baseline* & Deformable DETR & Deformable DETR & Yes & Yes & 46.4M & \textbf{0.291} \\
\midrule
\multirow{4}{*}{\cellcolor{white}RGB +Events} 
& FPN-fusion \cite{tomy2022fusing} & ResNet-50 & RetinalHead & No & No & / & 0.289 \\
& SFNet \cite{liu2024enhancing} & CSPDarknet53 & YOLOv5 & No & No & / & 0.412 \\
& SODFormer \cite{li2023sodformer} & Deformable DETR & Deformable DETR & Yes & No & 84.8M &  0.478 \\
\rowcolor{blue!10}
\cellcolor{white}& \algorithmname~(ours) & Deformable DETR & Deformable DETR & Yes & Yes & 87.9M & \textbf{0.506} \\
\bottomrule
\end{tabular}
}
\end{table*}

\begin{figure*}[t]
    \centering
    \includegraphics[width=\textwidth]{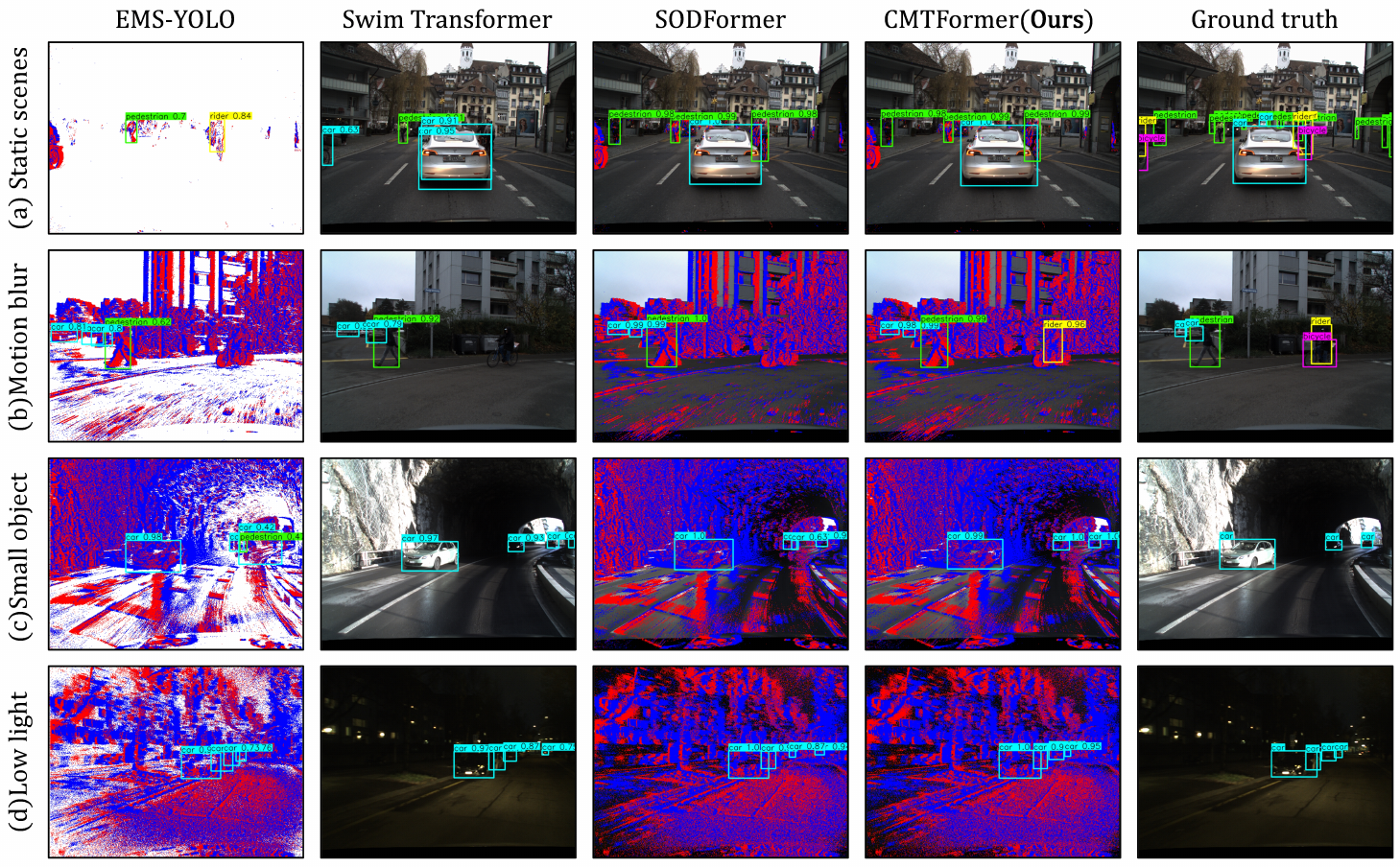}
    \caption{Visual comparison of \algorithmname~with state-of-the-art methods on the DSEC-Detection dataset~\cite{gehrig2024low}. From top to bottom: static scenes, high-speed motion blur, small object detection, and low-light scenes. Our method achieves superior detection performance across diverse scenarios.
    }
    \label{fig:various_scene_dsec}
    \vspace{-4mm}
\end{figure*}

We provide qualitative results on the DSEC-Detection dataset \cite{gehrig2024low}, as shown in Fig.~\ref{fig:various_scene_dsec}. From top to bottom, the four rows correspond to static scenes, high-speed motion blur, small object detection, and low-light environments. It can be observed that our \algorithmname, by leveraging the complementary characteristics of RGB frames and event data, maintains robust detection performance across diverse challenging conditions.
The third row of Fig.~\ref{fig:various_scene_dsec} further illustrates that \algorithmname~successfully detects distant and small vehicles, effectively addressing the failure cases commonly observed in single-modality detectors.

\subparagraph{Evaluation on PKU-DAVIS-SOD dataset.}
Table~\ref{tab:comparison2} provides a comprehensive comparison with ten representative detectors across three modality settings on the PKU-DAVIS-SOD dataset~\cite{li2023sodformer}. 

Event-driven methods rely only on asynchronous sparse event streams and generally struggle with motion noise. SSD-events~\cite{iacono2018towards}, NGA-events~\cite{hu2020learning}, YOLOv3-RGB reconstruction~\cite{redmon2018yolov3}) achieve only $0.221$, $0.232$ and $0.244$ mAP$_{50}$, respectively. More advanced temporal models, such as LSTM-SSD~\cite{liu2018mobile} and ASTMNet~\cite{li2022asynchronous}, provide moderate improvements up to $0.291$ mAP$_{50}$. Our event baseline, equipped with spatial-prior guided strategy and spatial-temporal attention, reaches $0.334$ mAP$_{50}$, outperforming Deformable DETR~\cite{zhu2020deformable} by $2.7$\% and ASTMNet~\cite{li2022asynchronous} by $4.3$\%.
These gains highlight the importance of spatial-prior guided strategy and temporal consistency for event-only recognition.

RGB-based methods benefit from richer textures but are vulnerable to motion blur and low-light conditions. Classical detectors such as Faster R-CNN~\cite{ren2016faster} and YOLOv3-RGB~\cite{redmon2018yolov3} achieve $0.443$ and $0.426$ mAP$_{50}$ respectively, while transformer-based detectors (Deformable DETR~\cite{zhu2020deformable} ($0.461$ mAP$_{50}$), LSTM-SSD~\cite{liu2018mobile} ($0.456$ mAP$_{50}$)) push the upper bound. However, our baseline achieves the best performance with $0.489$ mAP$_{50}$, surpassing Deformable DETR~\cite{zhu2020deformable} by $2.8$\%, LSTM-SSD~\cite{liu2018mobile} by $3.3$\%, Faster R-CNN~\cite{ren2016faster} by $4.6$\%. These improvements again demonstrate the benefit of spatial-prior guided spatial-temporal modeling in RGB-only conditions.

For the event-RGB object detection task, \algorithmname~achieves $0.525$ mAP$_{50}$, outperforming MFEPD~\cite{jiang2019mixed} by $8.7$\%, JDF~\cite{li2019event} by $8.3$\% and the previous state-of-the-art SODFormer~\cite{li2023sodformer} by $2.1$\%. 
\algorithmname~delivers consistently superior detection accuracy and robustness across challenging scenarios.

\begin{table*}[t!]
\centering
\caption{Comparison with state-of-the-art methods and our \algorithmname~on the PKU-DAVIS-SOD dataset\cite{li2023sodformer}. Methods marked with * employ the spatial-prior module.}
\label{tab:comparison2}
\resizebox{\textwidth}{!}{
\begin{tabular}{llllccc}
\toprule
Modality & Method & Input representation & Backbone & Temporal & Spatial-Prior & mAP$_{50}$ \\
\midrule
\multirow{7}{*}{\cellcolor{white}Event} 
& SSD-events \cite{iacono2018towards} & Event image & SSD & No & No & 0.221 \\
& NGA-events \cite{hu2020learning} & Voxel grid & YOLOv3 & No & No & 0.232 \\
& YOLOv3-RGB \cite{redmon2018yolov3} & Reconstructed image & YOLOv3 & No & No & 0.244 \\
& Faster R-CNN \cite{ren2016faster} & Event image & R-CNN & No & No & 0.251 \\
& Deformable DETR \cite{zhu2020deformable} & Event image & DETR & No & No & 0.307 \\
& LSTM-SSD \cite{liu2018mobile} & Event image & SSD & Yes & No & 0.273 \\
& ASTMNet \cite{li2022asynchronous} & Event embedding & Rec-Conv-SSD & Yes & No & 0.291 \\
\rowcolor{blue!10}
\cellcolor{white}& Our baseline* & Event image & Deformable DETR & Yes & Yes & \textbf{0.334} \\
\midrule
\multirow{4}{*}{\cellcolor{white}RGB} 
& Faster R-CNN \cite{ren2016faster} & RGB frame & R-CNN & No & No & 0.443 \\
& YOLOv3-RGB \cite{redmon2018yolov3} & RGB frame & YOLOv3 & No & No & 0.426 \\
& Deformable DETR \cite{zhu2020deformable} & RGB frame & DETR & No & No & 0.461 \\
& LSTM-SSD \cite{liu2018mobile} & RGB frame & SSD & Yes & No & 0.456 \\
\rowcolor{blue!10}
\cellcolor{white}& Our baseline* & RGB frame & Deformable DETR & Yes & Yes & \textbf{0.489} \\
\midrule
\multirow{4}{*}{\cellcolor{white}Events + RGB} 
& MFEPD \cite{jiang2019mixed} & Event image + RGB frame & YOLOv3 & No & No & 0.438 \\
& JDF \cite{li2019event} & Channel image + RGB frame & YOLOv3 & No & No & 0.442 \\
& SODFormer \cite{li2023sodformer} & Event image + RGB frame & Deformable DETR & Yes & No & 0.504 \\
\rowcolor{blue!10}
\cellcolor{white}& \algorithmname~(ours) & Event image + RGB frame & Deformable DETR & Yes & Yes & \textbf{0.525} \\

\bottomrule
\end{tabular}
}
\end{table*}

\subsection{Ablation Study}
\label{section:C}

\paragraph{Ablation of Each Component}
To investigate the contribution of each fusion stage and spatial-prior module, we take a standard feed-forward transformer detector (\ie, Deformable DETR\cite{zhu2020deformable} using only RGB frames) as our baseline. Based on this, we construct four variants by incrementally adding components from our proposed pipeline. As summarized in Table~\ref{table:ablation1}, the configurations are defined as follows:
(a) introduces LDFM, where features from RGB and event modalities are fused at the final stage;
(b) extends (a) by incorporating CEM, enabling deep semantic interaction;
(c) adds SAM on top of (b), allowing low-level cross-modality alignment;
Our \algorithmname~integrates all three fusion stages and spatial-prior module.
This design allows us to progressively validate the effectiveness of each module in terms of detection accuracy.

\begin{table}[t]
    \centering
    \caption{The ablation of each component, including SAM, CEM, LDFM, and the Spatial-Prior module. The baseline is Deformable DETR\cite{zhu2020deformable} using only RGB frames.}
    \resizebox{\columnwidth}{!}{
    \begin{tabular}{lccccc}
        \hline
        Method & SAM & CEM & LDFM & Spatial-Prior & mAP$_{50}$ \\
        \hline
        baseline &  &  &  &  &0.472  \\
        (a) &  &  & \checkmark &  & 0.478  \\
        (b) &  & \checkmark &\checkmark  &  & 0.484 \\
        (c) & \checkmark & \checkmark  & \checkmark &  &0.487  \\
        \rowcolor{blue!10}
        Ours & \checkmark & \checkmark & \checkmark & \checkmark &\textbf{0.506}  \\
        \hline
    \end{tabular}
    }
    \label{table:ablation1}
\end{table}

As shown in Table~\ref{table:ablation1}, using the late fusion strategy alone (a) improves mAP$_{50}$ by $0.6$\% compared to the baseline model, indicating that late fusion effectively integrates cross-modal features. However, relying only on late fusion does not fully exploit the complementarity between modalities.  
Adding the middle fusion module (b) further increases mAP$_{50}$ to $0.484$, indicating that middle fusion enhances deeper semantic interactions between modalities.
Building on this, incorporating the early fusion (c) raises mAP$_{50}$ to $0.487$, demonstrating that early fusion achieves low-level cross-modality alignment, further enhancing the model's discriminative ability.
Finally, \algorithmname~integrates all three fusion stages and introduces the spatial-prior mechanism, achieving an mAP$_{50}$ of $0.506$, which is a $3.4$\% improvement over the baseline model. This indicates that the collaborative effect of multi-stage fusion and spatial-prior mechanism significantly enhances the model's detection performance in complex scenarios.   
The results validate the effectiveness of our proposed method in enhancing spatial localization accuracy and modalities interaction, helping the model consistently focus on informative regions under challenging conditions.

\paragraph{Influence of Parameter in Spatial-Prior Module}
To evaluate the effect of the parameter \(\mu\) used in initializing object queries, we conduct a controlled experiment by varying \(\mu\) from $0.0$ (pure random queries) to $1.0$ (fully multimodal global context). As shown in Table~\ref{tab:ablation_param}, performance improves steadily as \(\mu\) increases, reaching a peak at \(\mu\) = $0.5$, where global prior and learnable queries are balanced. Further increasing \(\mu\) leads to slight performance degradation, possibly due to over-reliance on fused representations and reduced query diversity. This confirms that a moderate injection of multi-modal global context is beneficial.

\begin{table}[t]
\centering
\caption{The influence of the parameter $\mu$ in the spatial-prior module.}
\label{tab:ablation_param}
\resizebox{\columnwidth}{!}{
\begin{tabular}{c|cccccccc}
\toprule
Weight $\mu$ & 0.00 & 0.02 & 0.05 & 0.07 & 0.10 & 0.50 & 0.70 & 1.00 \\
\midrule
mAP         & 0.256 & 0.256 & 0.268 & \textbf{0.273} & 0.260 & 0.266 & 0.261 & 0.261 \\
mAP$_{50}$  & 0.487 & 0.485 & 0.498 & \textbf{0.506} & 0.490 & 0.484 & 0.489 & 0.485 \\
\bottomrule
\end{tabular}
}
\end{table}

\paragraph{Effect of Event Representations}
According to the results in Table~\ref{tab:event_representation}, using event images~\cite{maqueda2018event} as the input representation yields the best performance for \algorithmname~, achieving $0.506$ mAP$_{50}$, while maintaining inference speed comparable to other representations. The voxel grid representation~\cite{zhu2019unsupervised} performs slightly worse than event images but better than the Sigmoid representation~\cite{chen2018pseudo}. The Sigmoid representation~\cite{chen2018pseudo} suffers from a significant drop in detection performance due to the loss of some spatio-temporal information. These results indicate that event images better preserve the temporal and spatial characteristics of event streams, contributing to improved object detection accuracy. Conversely, a decline in representation quality directly impacts model performance. Therefore, selecting high-quality event representations is crucial for event-based vision detection in complex scenarios.

\begin{table}[t!]
    \centering
    \caption{Comparison of different event representations as input to \algorithmname.}
    \resizebox{\columnwidth}{!}{
    \begin{tabular}{lcc}
        \toprule
        Method & mAP$_{50}$ & Runtime (ms) \\
        \midrule
        Sigmoid representation~\cite{chen2018pseudo} & 0.471 & 53.8 \\
        Voxel grid~\cite{zhu2019unsupervised}        & 0.492 & 55.1 \\
        \rowcolor{blue!10}
        Event images~\cite{maqueda2018event}         & \textbf{0.506} & 54.6 \\
        \bottomrule
    \end{tabular}
    }
    \label{tab:event_representation}
\end{table}

\subsection{Scalability Test}
\label{section:D}
This section demonstrates that our \algorithmname~supports asynchronous processing of two modal visual streams and provides grad-cam visualizations of the progressive fusion strategy.

\begin{figure}[t!]
    \centering
    \includegraphics[width=\linewidth]{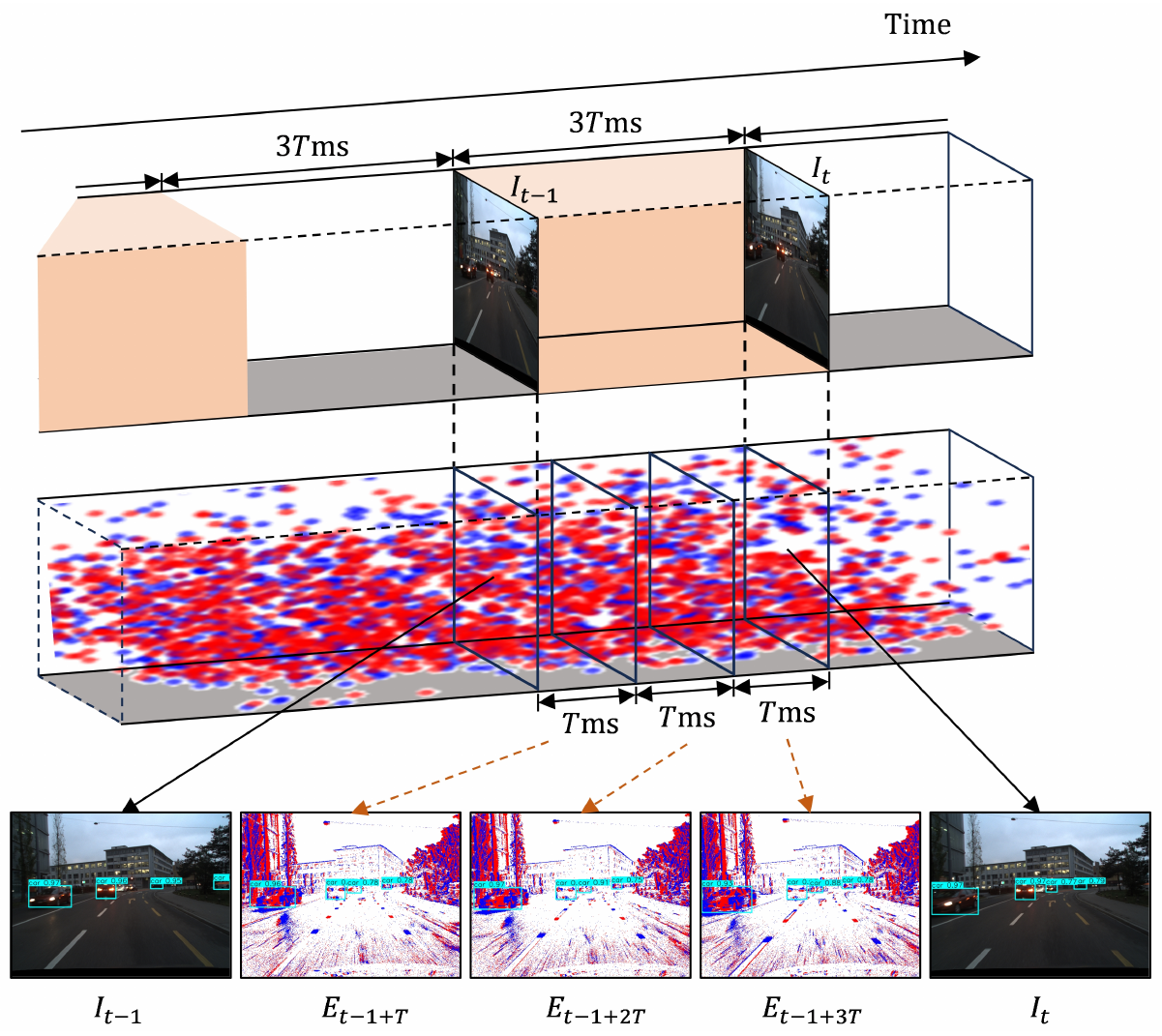}
    \caption{Visualization of asynchronous detections produced by combining RGB frames with the continuous event stream. The top row illustrates two adjacent RGB frames separated by a fixed temporal interval, while the middle row shows the stacked event slices captured at much finer temporal intervals. The bottom row presents asynchronous detection results corresponding to the two adjacent frames and the intermediate event stream.}
    \label{fig:asynchronous_inference}
\end{figure}

\paragraph{Asynchronous Inference Validation} 
Conventional global shutter cameras produce frames at limited rates, resulting in temporal gaps between adjacent frames and preventing truly continuous detection. Our \algorithmname~addresses this limitation by leveraging the high temporal resolution of event streams to supplement frame information between intervals. 
Specifically, we sample the event stream using a sliding window of 0.05 seconds with a step size of 0.0125 seconds, generating event data at 80 Hz. By combining 20 Hz frame data with 80 Hz event slices as input and achieve object detection output at 80 Hz. As shown in Fig.~\ref{fig:asynchronous_inference}, the predicted bounding boxes transition smoothly between frames, demonstrating that our approach enables continuous object detection.

\section{Conclusion}
In this work, we introduce \algorithmname, which marries transformer with hierarchical information interaction for RGB-event object detection. Our \algorithmname~integrates RGB frames and event streams through a hierarchical cross-modal fusion strategy, consisting of shallow-to-deep stages that enable hierarchical and purposive interaction across modalities. In addition, we devise a Spatial-Prior Module, which exploits global spatial context to adaptively steer the detector towards semantically and structurally informative regions, enhancing the localization effectiveness.
Extensive experiments on two challenging benchmarks verify that our \algorithmname~achieves state-of-the-art performance under diverse and adverse conditions. We hope the proposed framework can provide a robust and generalizable solution to pave the way for real-world multi-modal object detection.

\noindent \textbf{Declaration of competing interest}

The authors declare that they have no known competing financial interests or personal relationships that could have appeared to influence the work reported in this paper.

\noindent \textbf{Declaration of generative AI usage}

The authors do not use generative AI and AI-assisted technologies in the manuscript preparation process.

\noindent \textbf{Data availability}

The experiments are performed on popular benchmarks and the dataset samples can be accessed through public websites. Experimental data will be made available on request.

\noindent \textbf{Credit authorship contribution statement}

\textbf{Yu Li}: Conceptualization; Data curation; Methodology; Investigation; Software; Visualization; Validation; Writing – original draft; Writing – review \& editing; Formal analysis. \textbf{Yuenan Hou}: Supervision; Writing – original draft; Methodology; Writing – review \& editing; Project administration; Conceptualization. \textbf{Yingmei Wei}: Project administration; Funding acquisition; Supervision; Methodology; Writing – original draft. \textbf{Jiangming Chen}: Methodology; Writing – review \& editing; Supervision. \textbf{Yanming Guo}: Project administration; Supervision; Funding acquisition; Writing – review \& editing.

\noindent \textbf{Acknowledgment}

This work was supported by the National Natural Science Foundation of China (General Program, Grant No. 72571281), the Excellent Young Scientist Fund of Hunan Province (Grant No. 2025JJ40066).








\bibliographystyle{elsarticle-num}
\bibliography{references}

\end{document}